\documentclass[journal]{IEEEtran}
\ifCLASSINFOpdf
\else
\fi
\usepackage{graphicx}
\usepackage{blindtext}
\usepackage{amsmath}
\usepackage{amsfonts}
\usepackage{booktabs}
\usepackage{multirow}
\usepackage{multirow}
\usepackage{makecell}
\usepackage[scientific-notation=true]{siunitx}
\usepackage{pgf}
\usepackage{balance}
\usepackage{pgfpages}
\usepackage{hyperref}
\usepackage{color}
\usepackage{soul} 
\usepackage{todonotes}

\begin{document}
%
\title{Dynamic Emotion Modeling with Learnable Graphs and Graph Inception Network}
%
%
%

\author{Amir~Shirian,
        Subarna~Tripathi,~\IEEEmembership{Member,~IEEE,}
        and~Tanaya~Guha,~\IEEEmembership{Member,~IEEE}
\thanks{A. Shirian and T. Guha are with the Department
of Computer Science, University of Warwick, Coventry, UK.}
\thanks{S. Tripathi is with Intel Labs, San Diego, US.}
\thanks{}}

%
%

\markboth{IEEE Transactions on Multimedia,~Vol.~, No.~, February~2021}%
{Shirian \MakeLowercase{\textit{et al.}}: Bare Demo of IEEEtran.cls for IEEE Journals}
%



\maketitle

\begin{abstract}
Human emotion is expressed, perceived and captured using a variety of dynamic data modalities, such as speech (verbal), videos (facial expressions) and motion sensors (body gestures). We propose a generalized approach to emotion recognition that can adapt across modalities by modeling dynamic data as structured graphs. The motivation behind the graph approach is to build compact models without compromising on performance. To alleviate the problem of optimal graph construction, we cast this as a joint graph learning and classification task. To this end, we present the Learnable Graph Inception Network (L-GrIN) that jointly learns to recognize emotion and to identify the underlying graph structure in the dynamic data. Our architecture comprises multiple novel components: a new graph convolution operation, a graph inception layer, learnable adjacency, and a learnable pooling function that yields a graph-level embedding. We evaluate the proposed architecture on five benchmark emotion recognition databases spanning three different modalities (video, audio, motion capture), where each database captures one of the following emotional cues: facial expressions, speech and body gestures. We achieve state-of-the-art performance on all five databases outperforming several competitive baselines and relevant existing methods. Our graph architecture shows superior performance with significantly fewer parameters (compared to convolutional or recurrent neural networks) promising its applicability to resource-constrained devices. Our code is available at \href{https://github.com/AmirSh15/graph_emotion_recognition} {\texttt{/github.com/AmirSh15/graph\_emotion\_recognition}}.
\end{abstract}

%
\IEEEpeerreviewmaketitle

\section{Introduction}
\label{sec:intro}
Human emotion is expressed, perceived and captured using a variety of dynamic data modalities, such as speech (verbal), videos (facial expressions) and motion capture (body gestures). Modeling and analysis of these cues are critical for many human-centric systems with applications ranging from driver's safety to mental healthcare to human-robot conversational systems. In recent years, significant progress has been made towards the recognition and analysis of emotion using dynamic facial expressions \cite{li2017multimodal, pan2019deep}, speech \cite{parthasarathy2020semi, han2018towards} and body gestures \cite{crenn2017toward}.
Since human emotion is inherently multimodal, research efforts that combine information from multiple modalities are also on the rise \cite{ma2019audio}. Besides expressed emotion, work has also been done to analyze emotion evoked by natural images \mbox{\cite{zhang2019exploring}}, videos \mbox{\cite{zhang2018recognition}} and music \mbox{\cite{verma2019learning}}. 
\vspace{-0.1cm}
\par 
In the literature of dynamic emotion recognition, recurrent models, such as Long Short Term Memory networks (LSTM) are common \cite{han2018towards, lee2019visual}. These networks often have complex architecture with millions of trainable parameters requiring large amounts of training data. This makes many emotion recognition models incompatible for use in resource-constrained devices. A compact, efficient and scalable way to represent data is in the form of graphs. We thus adopt a graph approach to building a compact model for dynamic emotion recognition. Furthermore, existing emotion recognition models assume a prior knowledge of the input modality. Since emotion can be sensed through a variety of modalities, a generalized model that can handle disparate modalities efficiently is important. We show that our \emph{modality-agnostic} graph approach is able to achieve state-of-the-art accuracy across various modalities with significantly fewer trainable parameters. 
\begin{figure}[t]
\begin{center}
\includegraphics[width=1\linewidth, trim={0mm 1.3cm 0mm 0mm}, clip=true]{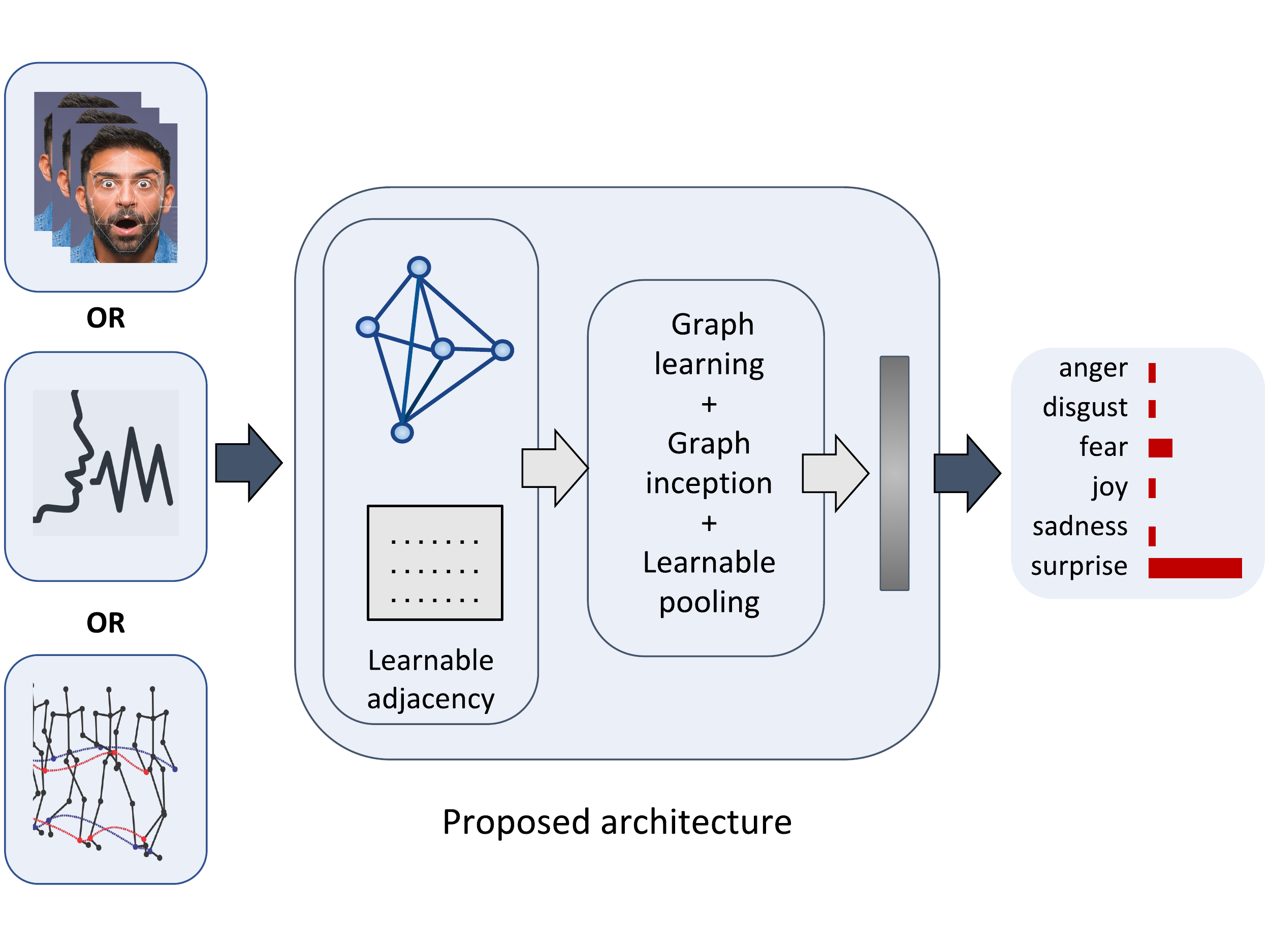}
\end{center}
\vspace{-0.2cm}
   \caption{A generalized graph approach to modeling emotion dynamics. Data samples are transformed to a \emph{learnable} graph structure, where each node corresponds to a short temporal segment or frame. A novel graph architecture (L-GRIN) produces an embedding for the entire graph facilitating emotion recognition.}
\label{fig:teaser}
\end{figure}
\par Traditionally, sequences are modeled using Recurrent Neural Networks (RNNs). However, recent literature has successfully used the idea of defining a sequence over a graph \cite{mao2018hierarchical,seo2018structured, yan2018spatial}. Considering a video frame sequence as a `structured' graph, Mao et al. showed that graph models can outperform RNNs \cite{mao2018hierarchical}. Motivated by these recent successes and in the pursuit of a compact model, we propose to adopt a graph approach to model emotion dynamics. Subsequently, we cast emotion recognition as a joint graph learning and classification problem (see Fig.~\ref{fig:teaser} for an overview). In our approach, each dynamic data sample is represented as a graph, where each node corresponds to a short temporal segment in the data. Each node is associated with the features extracted from the short temporal segment (frame) as its node attributes. This frame-to-node graph construction approach focuses on modeling the temporal dynamics in data; note that spatio-temporal structure (e.g., facial keypoints structure) within the graph resists the idea of a generic, modality-agnostic model and also increases model size significantly. Our graph structure (and hence the model) does not change with the choice of modality or node attributes. Modeling as a graph offers compactness and convenience to handle missing data (particularly common in mocap).

\par The graph structure i.e., the edge weights connecting the nodes is not naturally defined here. When a graph structure is not known apriori, a common practice is to manually construct the graph. This, however, results into sub-optimal graphs. We thus propose to learn the graph structure itself during the training stage. This is a generalized formulation, where the temporal dependencies between the nodes are automatically discovered. The only assumption we make is that the graph structure remains the same for all samples in a given database. To this end, we propose a novel Graph Convolution Network (GCN) architecture, the \emph{Learnable Graph Inception Network} (L-GrIN), with several novel components: a new definition of graph convolution that uses a non-linear layer-wise projection technique, introduction of an inception module in graph domain, learnable graph structure and a learnable graph-to-vector pooling function. Our architecture produces superior results on five benchmark emotion recognition databases spanning three different modalities (video, audio, mocap). Each database captures one of the following emotional cues: facial expressions, speech and body gestures. In summary, the main contributions of this paper are as follows:
\begin{itemize}
    \item A generalized, modality-agnostic graph approach to classify dynamic signals that combines graph learning with graph classification.
    \item A novel graph architecture, termed \textbf{L-GrIN}, with a new graph convolution layer, a graph inception module, learnable graph structure and learnable graph-to-vector pooling.
    \item State-of-the-art performance on dynamic emotion recognition tasks spanning three sensory modalities (video, audio, motion sensors) on five benchmark databases.
\end{itemize}
\section{Related Work}
\label{sec:related}
\begin{figure*}
\begin{center}
\includegraphics[width=1\linewidth, trim = 0cm 0cm 2.5mm 0cm, clip=true]{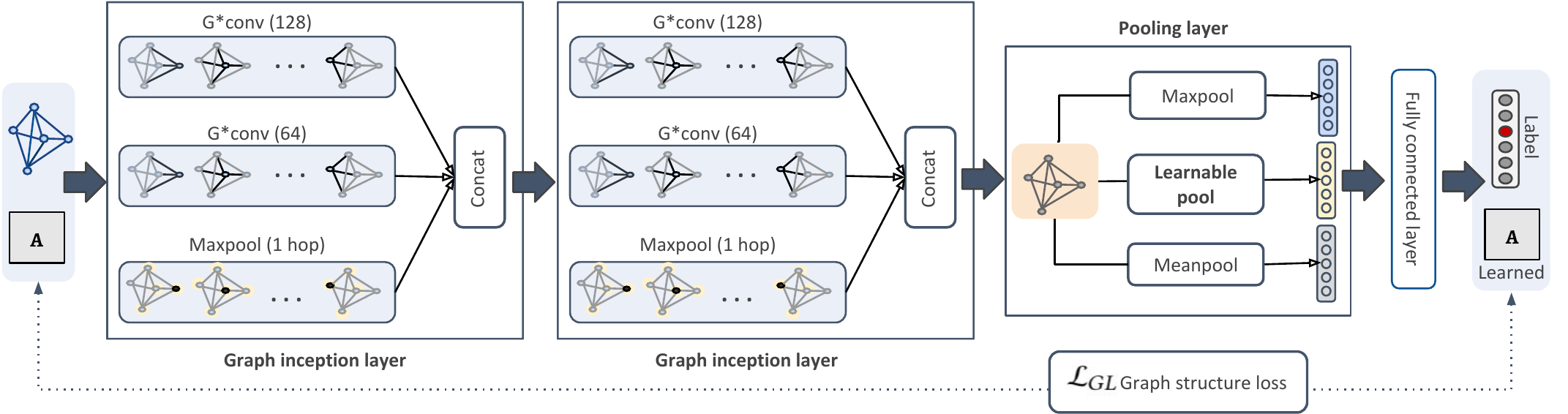}
\end{center}
    \caption{Our proposed architecture, L-GrIN, consists of two graph inception layers (with a new spectral graph convolution layer) and a pooling layer (two fixed pooling layers and a learnable pooling layer). The inception layers produce node-level representations that are pooled to obtain a graph-level representation by the pooling layer. L-GRIN also learns the underlying graph structure (adjacency matrix) by jointly optimizing a classification loss and a graph structure loss.}
\label{fig:overview}
\end{figure*}
In this section, we review the related work in the areas of GCNs and emotion recognition using various modalities.
\subsection{Graph neural network.}
Deep learning on graph data has emerged as a major topic in the past few years. This is because graphs provide a natural and convenient way to deal with large data. Among the different graph neural networks, GCNs have received the most attention \cite{gilmer2017neural,kipf2017semi, seo2018structured}. 
GCNs have been successfully applied to various image and video-based tasks, such as face clustering \cite{wang2019linkage}, object detection \cite{pointnet18}, and video representation learning \cite{mao2018hierarchical}. 
GCNs have been used to address skeleton-based action recognition recorded using motion capture \cite{yan2018spatial}. The application of graph networks has also started emerging in automatic speech recognition \cite{liu2016graph}.
\par
GCNs can be broadly classified into two categories: \emph{spatial} and \emph{spectral}. Spatial GCNs imitate the convolution operation of the Convolutional Neural Networks (CNN) by aggregating the information from neighboring nodes \cite{gilmer2017neural, niepert2016learning}. The problem of different graph nodes having different number of neighbours is usually addressed by using a fixed size neighborhood \cite{niepert2016learning} or by converting graph structures to a regular grid and subsequently applying traditional CNNs \cite{hamilton2017inductive}. A recent work proposed to develop the graph structure considering the Weisfeiler-Lehman graph isomorphism test \cite{xu2018graph}, and achieved state-of-the-art performance in node classification task in citation networks. On the other hand, spectral GCNs formulate the convolution operation as a frequency domain filtering operation following the theory of signal processing \cite{bruna2013spectral}, where convolution filters are seen as a set of learnable parameters. The ChebNet \cite{defferrard2016convolutional} is proposed to reduce the computational cost of spectral GCNs that redefined the convolution filter in terms of Chebyshev polynomials bypassing the need for eigen decomposition of the graph Laplacian. In a follow-up work \cite{kipf2017semi}, a first order approximation of the Chebyshev polynomials was introduced. This further simplified the spectral GCN computation as the convolution operation reduces to a linear projection.
\vspace{-1mm}
\subsection{Emotion recognition.} 
\subsubsection*{Facial emotion recognition.} Recognizing facial expressions is the most common way of analyzing emotion. The majority of work rely on identifying an individual's facial expression from images or videos (fewer work on videos), and associating them to one of the basic emotion classes. Recent efforts in image-based recognition are focused on using CNNs and its variants \cite{vemulapalli2019compact, hu2018squeeze}, and on using adversarial learning \cite{pan2019occluded}. A few works have proposed to use attention networks to account for the context \cite{Lee_2019_ICCV, marrero2019feratt, II_2019_CVPR}. 
RNNs and 3D CNNs have been used for video-based emotion recognition due to their ability to capture the temporal information \cite{jaiswal2016deep, fan2016video}. 
Another line of work focuses on the dynamics of landmark points in faces extracted from videos. In this context, a deep temporal appearance geometry network has been proposed \cite{jung2015joint} that uses the landmark point geometry and a CNN for emotion recognition. Another recent work constructed a trajectory matrix from the landmark points and used them as inputs to a CNN \cite{yan2018multi}.
\subsubsection*{Speech emotion recognition} 
Speech emotion recognition, especially using categorical labels, has been studied widely in the past years. Many speech emotion recognition systems still rely on low-level acoustic, prosodic and lexical features, that are then fed to deep models for classification. Other approaches use spectrograms (usually used as inputs to CNN models) \mbox{\cite{huang2014speech}} and even raw speech \mbox{\cite{latif2019direct}}. Recurrent models are prevalent due to their ability to capture the temporal dynamics of emotion \mbox{\cite{mirsamadi2017automatic,latif2019direct}}. A 3D RNN model has been recently proposed for end-to-end modeling \mbox{\cite{peng2018auditory}}. Attention-based techniques have been widely explored \mbox{\cite{mirsamadi2017automatic}}, \mbox{\cite{huang2017deep, gu2019mutual}}, while transformer-based architectures are gaining momentum in this field \mbox{\cite{tarantino2019self}}.
%
\subsubsection*{Body emotion recognition.} Body expressions are relatively less studied in emotion recognition.
The existing literature is focused on using motion information in terms of low-level descriptors, such as joint angles, 3D positions, distance between joints, velocity and acceleration \cite{crenn2017toward,DBLP:journals/corr/abs-1906-11884,DBLP:journals/corr/abs-1910-12906}. A trajectory learning approach \cite{crenn2017toward} proposed to identify `neutral' motion from input data, and used the deviation of a given input from the neutral motion as a feature for classifying emotions. Another recent work combined deep features with psychological attributes to detect emotion from 3D body pose using a random Forest classifier \cite{DBLP:journals/corr/abs-1906-11884}. Gait information has also been considered for recognizing emotion, where a spatial GCN is used to detect the emotional state \cite{DBLP:journals/corr/abs-1910-12906}.
%
\section{Proposed Approach}
\label{sec:proposed}
In this section, we describe our deep graph approach to emotion recognition. First, we construct a graph from dynamic input data following a generalized frame-to-node approach. Next, we propose a novel architecture that jointly performs graph learning and graph classification. This is achieved by optimizing over a new loss function that combines classification loss and a graph structure loss. The proposed architecture, L-GRIN, is illustrated in Fig.~\ref{fig:overview}. Below, we describe each component of this network in detail.
\vspace{-2mm}
\subsection{Graph construction} 
\label{subsec:graphcon}
Given a dynamic input sequence, our first task is to construct an undirected graph $\mathcal{G} = (\mathcal{V},\mathcal{E})$ that can efficiently capture the emotion-related dynamics in the data, where $\mathcal{V}$ is the set of nodes with cardinality $\vert\mathcal{V}\vert = M$ and $\mathcal{E}$ is the set of all edges between the connected nodes. A representative description of $\mathcal{G}$ is typically given by an adjacency matrix $\mathbf{A}\in\mathbb{R}^{M\times M}$ which is symmetric for an undirected graph. 

\begin{figure}[t]
\begin{center}
   \includegraphics[width=0.7\linewidth, trim=5cm 2.3cm 4.6cm 3mm, clip=true]{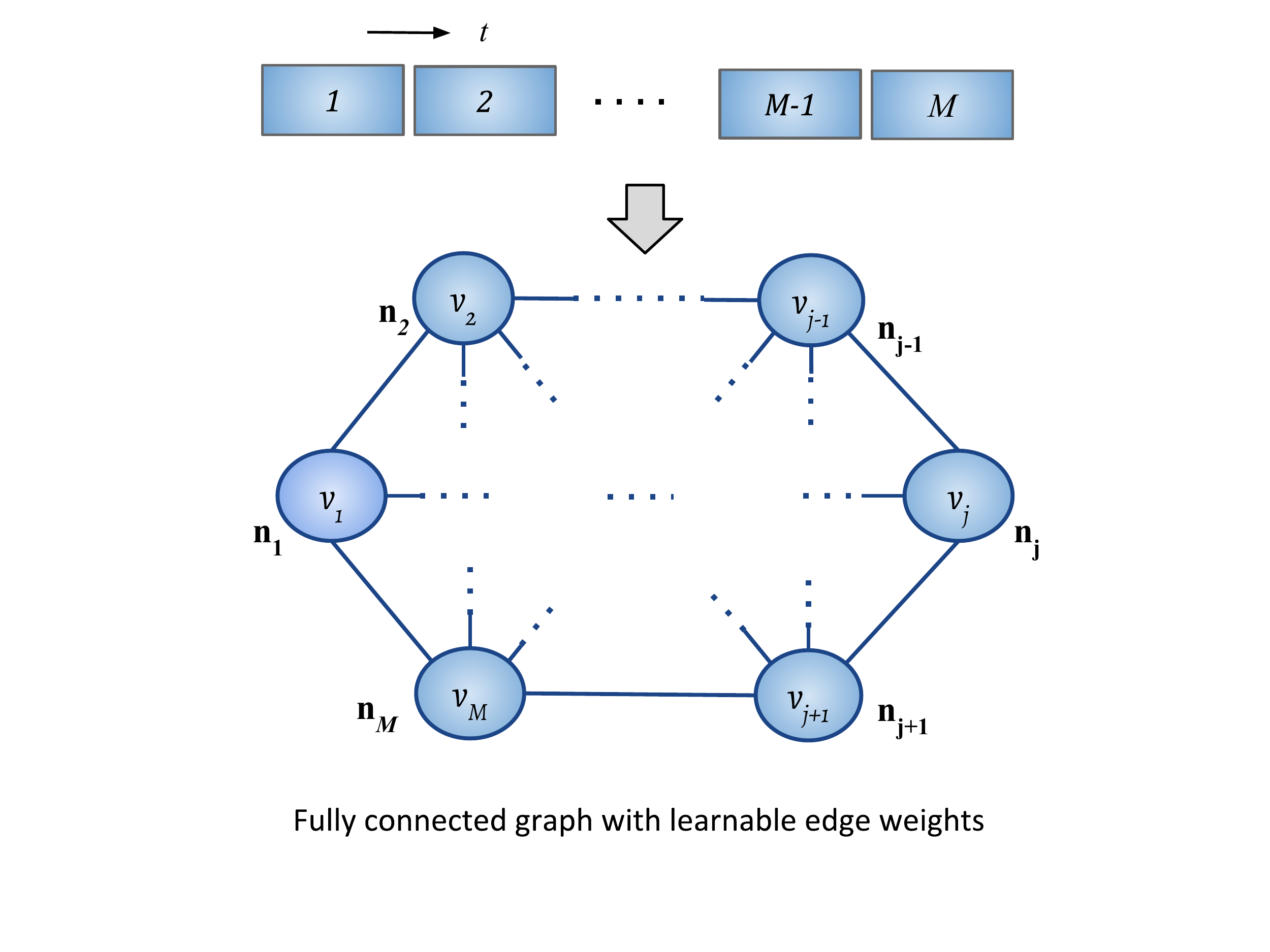}
\end{center}
\vspace{-0.2cm}
   \caption{Graph construction: Given a dynamic input sequence of $M$ segments, a fully-connected graph with $M$ nodes is constructed without making any assumption. The edge weights are learned during the training phase. Each node is associated with a node attribute vector $\mathbf{n}_i$.}
   \label{fig:graph_con}
\end{figure}

Our graph construction approach follows a \emph{frame-to-node} transformation, where $M$ frames in the data form the $M$ graph nodes $\{v_i\}_{i=1}^M \in \mathcal{V}$ (see Fig.~\ref{fig:graph_con}). A frame refers to a small temporal segment of the data, e.g., an audio segment of length 40ms. In order to encode the temporal information, a frame (node) should be connected with weights to a series of past and future nodes. An element $(\mathbf{A})_{ij} \in \mathbf{A}$ contains the weight corresponding to the edge $e_{ij}\in \mathcal{E}$ connecting $v_i$ and $v_j$. Note that the graph structure is not naturally defined here, i.e., the elements in $\mathbf{A}$ are unknown. A common way to define the elements in $\mathbf{A}$ is through constructing a distance function manually \cite{yan2018spatial}. However, this may result into a sub-optimal graph representation. Hence, we propose to learn the elements in $\mathbf{A}$ by jointly optimizing a structural loss combined with a classification loss. This loss function will be discussed in Section \ref{subsec:joint}.

In order to capture the emotion content at node level, we rely on modality-specific features or even, raw data. Each node $v_i$ is thus associated with a \textit{node feature} vector $\mathbf{n}_i \in \mathbb{R}^{P}$. 
A feature matrix ${\bf N}\in\mathbb{R}^{M\times P} $ consisting all the node feature vectors is defined as ${\bf N}= [\mathbf{n}_1, \mathbf{n}_2, \cdots, \mathbf{n}_M]^T$.
Features for individual modalities is discussed in Section \ref{sec:Exp}.
\subsection{Learnable graph inception network}
\label{subsec:joint}
Given a set of (dynamic inputs transformed to) graphs $\{G_1, ..., G_N\}$ and their true labels $\{\mathbf{y}_1, ..., \mathbf{y}_N\}$, our task is to develop a deep graph architecture that is able to recognize the emotional content in the data. Since the graph structure is not naturally defined here, we also learn an optimal $\mathbf{A}$ from the training data with the underlying assumption that each graph has different node features but the same edge weights. We formulate this as a joint graph learning and graph classification problem.

A common GCN architecture takes the node feature matrix $\mathbf{N} \in\mathbb{R}^{M\times P}$ and the graph adjacency matrix $\mathbf{A}$ as inputs and produces a \emph{node-level} representation matrix $\mathbf{Z}\in\mathbb{R}^{M \times Q}$, where $Q$ is the dimension of the output feature vector at each node. A GCN layer $\mathbf{H}^{(k+1)}$  can be defined as a non-linear function of its previous layer as follows
\begin{equation}
    {\bf H}^{(k+1)} = \sigma({\bf AH}^{(k)}\mathbf{{W}}^{(k)})
\end{equation}
where $\mathbf{W}^{(k)}$ is the weight matrix for the $k^{th}$ layer of the neural network, $\sigma$ is a non-linear activation function, such as a ReLU, and $k$ is the layer number ($k = 0, \cdots K$). Note that $\mathbf{H}^{(0)} = \mathbf{N}$ and $\mathbf{H}^{(K)} = \mathbf{Z}$.
An effective improvement on this propagation rule has been recently proposed \cite{kipf2017semi}.
\begin{equation}
    {\bf H}^{(k+1)} = \sigma(\mathbf{D}^{-\frac{1}{2}} (\mathbf{A +I})\mathbf{D}^{-\frac{1}{2}}\mathbf{H}^{(k)}\mathbf{W}^{(k)})
    \label{eq:gcn}
\end{equation}
where $\mathbf{D}$ is the degree matrix of $\mathbf{A}$, and $\mathbf{I}$ is an $M\times M$ identity matrix. Note that the terms within the parenthesis in Eq.~\eqref{eq:gcn} is simply a linear projection, and can be re-written as
\begin{equation} \label{eq:gcn_simple}
    {\bf H}^{(k+1)} = \sigma( \mathbf{\hat{A}}\mathbf{H}^{(k)}\mathbf{W}^{(k)})
\end{equation}
where $\mathbf{\hat{A}}= \mathbf{D}^{-\frac{1}{2}} (\mathbf{A +I})\mathbf{D}^{-\frac{1}{2}}$.
\par We present a new GCN architecture, called  L-GrIN (see Fig.~\ref{fig:overview}), for joint graph learning and classification. It has the following four new components:
\par
\vspace{3mm}\noindent
$\bullet\,\,$ \textbf{Non-linear spectral graph convolution ({\bf $\mathcal{G}^*$conv}).} Motivated by a recent work on spatial graph neural network \cite{xu2018how}, we replace the linear projection in \eqref{eq:gcn_simple} by a multi-layer perceptron (MLP) layer, and replace $\hat{\mathbf{A}}$ by a learnable $\mathbf{A}$. Thus, instead of the linear layer in \eqref{eq:gcn_simple}, we define a new spectral graph convolution layer $\mathcal{G}^*(\cdot)$ as follows:
\begin{align} \label{eq:gcn_MLP}
    \mathcal{G}^*(\mathbf{H}^{(k)}) &= 
    \sigma\Big(\mathrm{MLP}^{(k)}\big(\mathrm{ReLU}(\mathbf{A})\mathbf{H}^{(k)}\big)
    \Big)
\end{align}
where $\mathrm{MLP}(.)$ has two hidden layers with $\eta$ neurons each, $\mathbf{A}$ is the learnable adjacency matrix and $\sigma$ is a nonlinear activation function. $\mathbf{A}$ is learned through a joint optimization process described later in this section. The $\mathrm{ReLU}(\cdot)$ in Eq.~\eqref{eq:gcn_MLP} ensures the non-negativity of $\mathbf{A}$. We refer to the convolution operation defined above as {\bf $\mathcal{G}^*$conv} in the rest of the paper.
\par \vspace{2mm}\noindent
$\bullet\,\,$ \textbf{Graph inception.}
We extend the idea of inception layer in traditional CNNs \cite{szegedy2015going} to the graph domain, and introduce a \emph{graph inception} module in our architecture (see Fig.~\ref{fig:overview}). Our graph inception layer consists of two graph convolution layers and one maxpool layer operating on directly connected (1-hop) neighbours only. 

Given an input $\mathbf{H}^{(k)}$, the proposed graph inception layer is defined as follows:
\begin{equation}
\label{eq:incept}
    {\bf H}^{(k+1)} = \Big [\mathcal{G}^*_1(\mathbf{H}^{(k)})\, \vert \, \mathcal{G}^*_2(\mathbf{H}^{(k)})\, \vert \,
    \mathrm{maxpool}(\mathbf{H}^{(k)})\Big ]
\end{equation}
\noindent where $\vert$ denotes concatenation of the node features, and $\mathcal{G}^*_1$ and $\mathcal{G}^*_2$ are two {\bf $\mathcal{G}^*$conv} layers (see Eq.~\eqref{eq:gcn_MLP}) with different size of their MLP layers ($\eta =128$ for $\mathcal{G}^*_1$ and $\eta = 64$ for $\mathcal{G}^*_2$). Hence, for an input of $\mathbf{H}^{(k)}\in\mathbb{R}^{M \times P}$, the inception layer produces an output $\mathbf{H}^{(k+1)}\in\mathbb{R}^{M \times (128+64+P)}$. 

The motivation behind the inception layer is to be able to capture the emotion dynamics at multiple temporal scales. The two {\bf $\mathcal{G}^*$conv} layers that yield embeddings of different dimensions can be interpreted as a \emph{multiscale analysis} on graphs in spectral domain. Like a traditional inception layer in CNN, our graph inception layer also combines features from multiple scales allowing the network to have both width and depth. Our graph inception layer has fewer parameters (compared to inception networks in CNNs) enabling us to feed the input directly to the inception layer.

The maxpool function in Eq.~\eqref{eq:incept} operates on every node separately. For each node $v_i$, we only consider its directly connected neighbors (1-hop), and maxpool over the embeddings along feature dimension. Note that as we start with a fully-connected graph, initially this operation is the same as maxpooling over all nodes, but this changes quickly as we start learning the graph structure. 
\par
\vspace{2mm}
\noindent
$\bullet\,\,$ \textbf{Learnable pooling for graph-level representation.} Our objective is to classify entire graphs, as opposed to the more common task of classifying each node. Hence, we seek a \emph{graph-level} representation $\mathbf{h}_G\in\mathbb{R}^Q$ as the output of our network. This can be obtained by pooling the node-level representations $\mathbf{H}^{(k)}$ at the $K$-th layer before passing them to the classification layer (see Fig.\ref{fig:overview}). Common choices for pooling functions in graph domain are mean, max and sum pooling. Max and mean pooling often can not preserve the underlying information about the graph structure while sum pooling is shown to be a better alternative \cite{xu2018how}. However, all these pooling functions treat every neighboring node with equal importance, which may not be optimal. To this end, we propose to \textit{learn} a pooling function $\Psi$ that combines the node embeddings from the $K$-th layer to produce an embedding for the entire graph. Additionally, we also use maxpool and meanpool and combine all the graph-level embeddings together. The pooling layer is thus defined as follows: 
\begin{equation} \label{eq:graph_emb}
    \mathbf{h}_G = \big[\mathrm{maxpool}(\mathbf{H}^{(K)})\,|\,\Psi(\mathbf{H}^{(K)})\,|\,\mathrm{meanpool}(\mathbf{H}^{(K)})\big] 
\end{equation}
    $$\Psi(\mathbf{H}^{(K)})= \mathbf{H}^{(K)} \mathbf{p}$$
where $\mathbf{p}$ has learnable weights to combine node-level embeddings to obtain a graph-level embedding. 
\par
\vspace{2mm}
\noindent
$\bullet\,\,$ \textbf{Learnable adjacency} ($\mathbf{A}$). Recall that in our task the graph structure is not known. Although we can define such structure manually, results are sub-optimal. An effective approach would be to learn the graph structure (adjacency matrix) itself by jointly optimizing over a classification loss and graph learning loss. We assume that all videos have the same underlying graph structure containing the same number of nodes and edges. This largely simplifies our task of graph structure learning. The overall loss $\mathcal{L}$ for joint graph learning and classification is composed of two components: 
(i) $\mathcal{L}_{GC}$: a graph classification loss, and (ii) $\mathcal{L}_{GL}$: a graph learning loss. The classification loss $\mathcal{L}_{GC}$ is defined as the cross-entropy loss:
\begin{equation} \label{eq:GCN-loss}
\begin{aligned}
    \mathcal{L}_{GC}=-\displaystyle\sum_{n=1}^{N} \mathbf{y}_n \log \mathbf{\hat{y}}_n
\end{aligned}
\end{equation}
where $\mathbf{\hat{y}}_n$ is the predicted label for the $n^{th}$ sample. The graph learning loss, $\mathcal{L}_{GL}$, is designed to facilitate learning the pooling vector $\mathbf{p}$ and the adjacency matrix $\mathbf{A }$. This is defined as follows:
\begin{equation} \label{eq:GL-loss}
\begin{aligned}
    \mathcal{L}_{GL}= \underbrace{\lambda_1 \mathbf{e}^{T}(\mathbf{A}_d \odot \mathbf{A})\mathbf{e} +
    \lambda_2 \Vert \mathbf{A} \Vert^2_F}_{\text{graph structure loss}} \,\,\,\, + 
    \underbrace{\lambda_3 \Vert \mathbf{p} \Vert^2_2}_{\text{learnable pooling}}
\end{aligned}
\end{equation}
where $\odot$ denotes element-wise multiplication, $\mathbf{e}$ is an all-ones vector, $\Vert \cdot \Vert_F$ denotes Frobenious norm, $\lambda_1$, $\lambda_2$, and $\lambda_3$ control the relative weights of the three terms, and $\mathbf{A}_d$ is a structure matrix defined as follows:
\begin{equation} \label{eq:structure}
    (\mathbf{A}_d)_{ij} = (i-j)^2
\end{equation}
The structure matrix $\mathbf{A}_d$ forces the nodes that are temporally close to each other to have stronger relationship, i.e.\ higher weights in the $\mathbf{A}$. The larger the squared distance between two nodes $v_i$ and $v_j$ (frames), the smaller will be the weights in $(\mathbf{A})_{ij}$. The ReLU operation (see Eq.~\eqref{eq:gcn_MLP}) ensures the non-negativity of the elements in $\mathbf{A}$.
The overall optimization is thus as follows: 
\begin{equation*}
    \underset{\mathbf{A}, \mathbf{p}, \mathbf{\Theta}^{(1:k)}}{\text{min }}\mathcal{L} \,= \underset{\mathbf{A}, \mathbf{p}, \mathbf{\Theta}^{(1:k)}}{\text{min }}{\big [\mathcal{L}_{GC} + \mathcal{L}_{GL}\big ]}
\end{equation*}
where, $\Theta$ denotes all other learnable network parameters across all graph convolution layers including its constituent MLP layers. Every term in the overall loss function $\mathcal{L}$ is differentiable, thereby allowing an end-to-end optimization.
%
\section{Experiments}
\label{sec:Exp}
We now present extensive experimental results and analysis to evaluate the performance of the proposed architecture for facial, speech and body emotion recognition. 
\subsection{Facial emotion recognition}
\subsubsection*{\bf Video databases} We use three large video emotion recognition databases for our experiments. The databases are chosen based on their popularity in emotion recognition literature.

\noindent
The \textbf{RML} database \cite{wang2008recognizing} contains 720 videos of 6 basic emotions: \emph{anger, disgust, fear, joy, sadness}, \emph{surprise} collected when the subjects speak. The subjects are from various ethnic groups and speak different languages.

\noindent The \textbf{eNTERFACE} \cite{martin2006enterface} is contains 1170 videos of 42 subjects with six basic emotion classes as RML. These emotions are the reactions after listening to six different short stories, where each person reads out 5 phrases based on their emotional reaction.

\noindent
The \textbf{RAVDESS} database \cite{livingstone2018ryerson} contains 4904 videos labeled with 8 classes: \emph{anger, calmness, disgust, fear, joy, neutral, sadness} and \emph{surprise}. This is the largest video emotion database currently available.

\subsubsection*{\bf Node features} 
\begin{table}[t]
\caption{Facial emotion recognition results on three video databases.}
\label{tab:all_baselines_faces}
\begin{center}
\renewcommand*{\arraystretch}{1.6}
\begin{tabular}{p{2.4cm}| c c c|c}
\hline
\bf \multirow{2}{*}{Model}  & \multicolumn{3}{c|}{\bf Accuracy (\%) } & \bf \multirow{2}{*}{Params}\\
\cline{2-4}
    & RML  & eNTERFACE & RAVDESS & \\
\hline \hline
*BLSTM & 60.00   &  58.67  & 56.14          & $\sim1$M\\
*GCN \cite{kipf2017semi}     &76.57 &  69.81 & 69.34        & $\sim102$K\\
*PATCHY-SAN \cite{niepert2016learning}  & 80.00 & 67.49 & 73.52  & $\sim 52$K\\ 
*PATCHY-Diff \cite{ying2018hierarchical}  & 85.59  & 76.96  &79.83 & $\sim 71$K\\
SENet \cite{hu2018squeeze}  &71.20  & 79.22 &71.06   & $\sim 26$M\\
AVEF \cite{ma2019audio} & 82.48 & 85.69 & - & -\\
KCFA \cite{wang2012kernel} & 82.22 & 76.00 & - & -\\
OKL \cite{seng2016combined} & 90.83 & 86.67 &- &  - \\
TJE \cite{ghaleb2019multimodal} & - & - & $72.30$ & -\\
*{\bf L-GrIN} & \bf{94.11} & \bf 87.49  & \bf{85.65} & $\sim 120$K \\
\hline
\multicolumn{3}{l}{*use same node features}
\end{tabular}
\end{center}
\end{table}
\begin{figure}[t]
 \centering
 \begin{minipage}[t]{0.44\linewidth}
    \centering
  \includegraphics[width=1\linewidth, height=0.98\linewidth, trim={30mm 12mm 31mm 15mm}, clip=true, keepaspectratio=false]{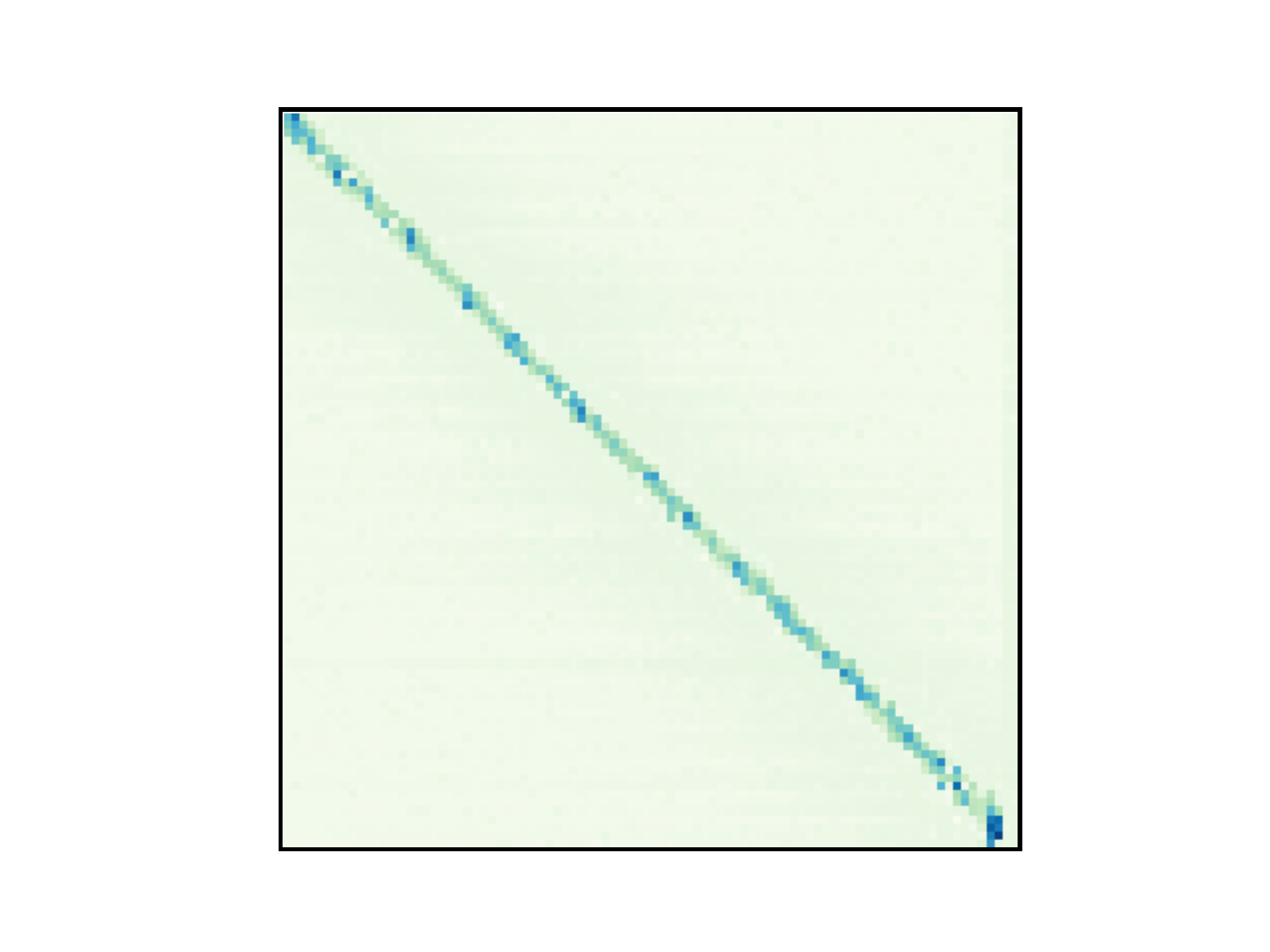}
    {\textbf{RAVDESS} ($90\times 90$)}
 \end{minipage}\hspace{1mm}
 \begin{minipage}[t]{0.44\linewidth}
    \centering
    \includegraphics[width=1.08\linewidth, height=1\linewidth, trim={26mm 12mm 22mm 13mm}, clip=true, keepaspectratio=false]{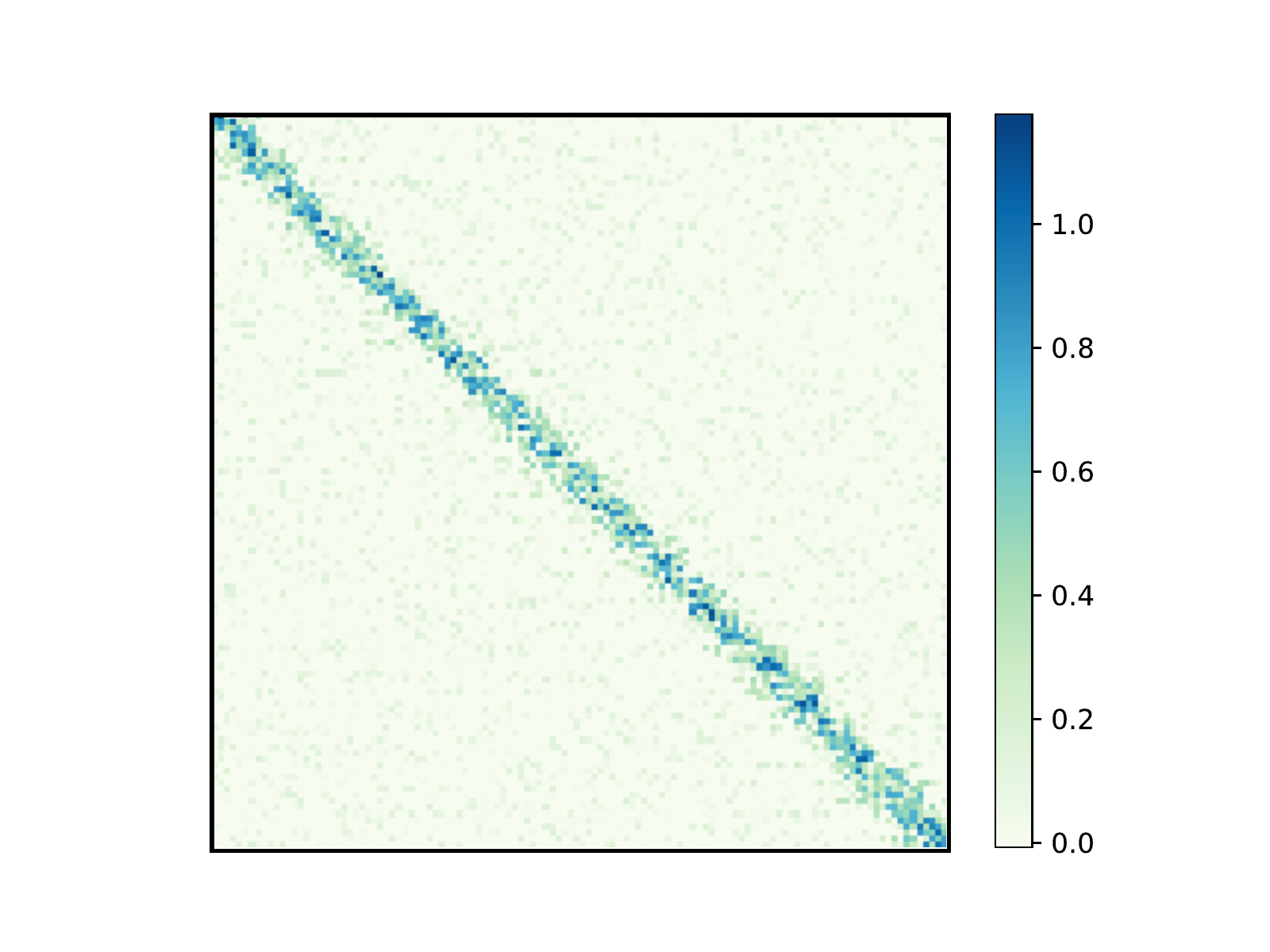}
    {\textbf{MPI} ($120\times 120$)}
 \end{minipage}
 \caption{\emph{Learned} adjacency matrices for facial and body emotion recognition showing strong temporal dependency between neighboring segments. Darker values indicate higher weights.}
 \label{fig:adj}
\end{figure}
The databases we use provide only raw video clips. We choose to use facial landmark points extracted from the video frames as node attributes. This is because landmark points are known to effectively capture the facial dynamics \cite{gu2017dynamic}. We extract $68$ landmark points at every video frame using a state-of-the-art landmark detection method \cite{bulat2017far}, resulting into node feature vectors of dimension $P = 136$.

\subsubsection*{\bf Implementation details} 
We use a 10-fold cross-validation for all three databases, and report the average recognition accuracy in Table \mbox{\ref{tab:all_baselines_faces}}.
We fix the length of each input video to $90$ frames yielding a graph with $M=90$ nodes. The shorter videos are simply padded by duplicating frames from the beginning of the video (cyclic padding). Our network weights are initialized following the Xavier initialization.
We set $\lambda_1 = \lambda_2 = 0.1$ and $\lambda_3=1 \times 10^{-4}$ (see Eq.~\ref{eq:GL-loss}). We used Adam optimizer with a learning rate of $0.01$ and decay rate of $0.5$ after each $50$ epochs for all experiments. To initialize the learnable adjacency matrix $\mathbf{A}$, we generate a random matrix whose elements are drawn from a Normal distribution with zero mean and unit variance. We used Pytorch for implementing our model and the baselines, and an NVIDIA RTX-2080Ti GPU for all experiments. 

\subsubsection*{\bf Baselines, state-of-the-art} We compare our model against two competitive and relevant baselines as follows: 

\noindent
\textbf{BLSTM.} The first baseline is a Bidirectional LSTM (BLSTM), an extension of the traditional LSTMs \cite{graves2005framewise,tan2016lstm}. LSTM and its variants have been successfully used in sentiment analysis in language and speech. 
This BLSTM comprises 1-layered bidirectional cells with embedding size $300$ followed by a fully connected layer.

\noindent
\textbf{GCN} \cite{kipf2017semi}. A natural baseline to compare with our model is a spectral GCN in its standard form (as in Eq.~\eqref{eq:gcn_simple}). The original network \cite{kipf2017semi} is designed for node classification and only yields node-level embeddings. To obtain a graph-level embedding, we used max and mean pooling at the end of convolution layers. The GCN uses a binary adjacency matrix constructed following the method used in graph-based action recognition \cite{yan2018spatial}. 

In addition to the baselines, we compare with two state-of-the-art graph classification architectures:

\noindent
\textbf{PATCHY-SAN} \cite{niepert2016learning} is a recent architecture that learns CNNs for arbitrary graphs. This architecture is originally  developed for  graph  classification.

\noindent
\textbf{PATCHY-Diff} \cite{ying2018hierarchical} is referred to an architecture where PATCHY-SAN is used in combination with the differentiable pooling layer between graph convolution layers proposed recently \cite{ying2018hierarchical}.

\noindent
\textbf{SENet} \cite{hu2018squeeze}, Squeeze and Excitation net is a state-of-the-art CNN architecture recently proposed for facial emotion recognition in videos.

Comparisons are also made with other existing works on the respective databases: AudioVisual Emotion Fusion (AVEF) \cite{ma2019audio}, Kernel Crossmodal Factor Analysis (KCFA) \cite{wang2012kernel}, Optimized Kernel-Laplacian (OKL) \cite{seng2016combined} and Temporal Joint Embeddings (TJE) \cite{ghaleb2019multimodal}.

\vspace{2mm}
\subsubsection*{\bf Results} 
Table \ref{tab:all_baselines_faces} compares the performance of L-GrIN with all the methods mentioned above. Clearly, the proposed model outperforms all the existing methods by a significant margin, including the graph-based state-of-the-art architectures, such as PATCHY-SAN and PATCHY-Diff. Our model performs better than BLSTM - a class of models most commonly used in video-based emotion recognition. SENet is a very recent CNN architecture developed for emotion recognition, which also trails our model in terms of performance. When compared to the GCN baseline \cite{kipf2017semi}, L-GrIN improves the recognition accuracy by more than 10$\%$ on RML and eNTERFACE, and more than 5$\%$ on RAVDESS. 

Also note that KCFA, OKL and TJE use both audio and visual information for recognition. Our model, even though uses only visual information, shows significant improvement over the audiovisual methods.

Fig.~\ref{fig:adj} shows the learned adjacency matrix for the RAVDESS database. The learned graph structure shows higher values closer to the diagonal i.e., the weights shared among the neighboring nodes. This indicates higher temporal dependencies locally and weaker dependency as we go further from a node. 
\subsection{Speech emotion recognition}
\begin{table}[t]
\caption{Speech emotion recognition results on IEMOCAP database.}
\label{tab:all_baselines_speech}
\vspace{-2mm}
\begin{center}
\renewcommand*{\arraystretch}{1.5}
\begin{tabular}{p{3cm}| c | c}
\hline 
\bf \multirow{1}{*}{Model}  & \multicolumn{1}{c|}{\bf Accuracy (\%)} & \bf \multirow{1}{*}{Params}\\
 \hline \hline
$^*$BLSTM (baseline) & $58.04$           & $\sim 0.8$M \\
$^*$GCN (baseline)     &   $56.14$    & $\sim 78$K \\
$^*$PATCHY-SAN \cite{niepert2016learning}  &  $60.34$ & $\sim 60$K\\ 
$^*$PATCHY-Diff \cite{ying2018hierarchical}  & $63.23$  & $\sim 68$K \\
CNN \cite{latif2019direct} & $58.52$ & $\sim 0.45$M\\
CNN-LSTM \cite{latif2019direct} & $59.23$ & $\sim 0.6$M \\
Rep learning \cite{ghosh2016representation} & $50.40$ & - \\
LSTM-CTC \cite{han2018towards} & $64.20$ & $\sim 0.4$M \\
$^*${\bf L-GrIN} &  $\mathbf{65.50}$ & $\sim 92$K \\
\hline
\multicolumn{3}{l}{$^*$ use same node features}
\end{tabular}
\end{center}
\end{table}
\subsubsection*{\bf Databases} We use the popular IEMOCAP database \cite{busso2008iemocap} for evaluating the performance of our model on speech emotion recognition. This database contains a total of 12 hours of data recorded in 5 sessions, where each session contains utterances from two speakers. The final database contains a total of 5531 utterances: 1103 \emph{angry}, 1708 \emph{neutral}, 1636 \emph{happy} and 1084 \emph{sad}.

\subsubsection*{\bf Node features} We extract a set of low-level descriptors (LLDs) from the raw speech utterances as proposed for Interspeech2009 emotion challenge \mbox{\cite{schuller2009interspeech}} using the OpenSMILE toolkit \mbox{\cite{eyben2009openear}}. The feature set includes Mel-Frequency Cepstral Coefficients (MFCCs), zero-crossing rate, voice probability, fundamental frequency (F0) and frame energy. For each sample, we use a sliding window of length 25ms with a stride length of 10ms to extract the LLDs locally. Each feature is then smoothed using a moving average filter, and the smoothed version is used to compute their respective first order delta coefficients. In addition, motivated by a recent work on speech emotion recognition \mbox{\cite{mangalam2018learning}}, we also add \emph{spontaneity} as a binary feature. The spontaneity information comes with the database. Altogether this produces node feature vectors of dimension $P=35$.

\subsubsection*{\bf Implementation details} Each audio sample produces a graph of $M=120$ nodes, where each node corresponds to a (overlapping) speech segment of length 25ms. Cyclic padding is used to make the samples of equal length, as before. We perform a 5-fold cross-validation and report the average unweighted accuracy in Table \ref{tab:all_baselines_speech}. The unweighted accuracy, a standard evaluation strategy for IEMOCAP, does not take into account the class imbalances. It simply computes the total number of correct classifications across all classes. All other parameters and settings remain the same as before to show the generalizability of our model.

\subsubsection*{\bf Baselines, state-of-the-art} Our model is compared with two baselines (BLSTM and GCN), two state-of-the-art graph-based architectures (PATCHY-SAN and PATCHY-Diff) as before. In addition, we also compare our model with four state-of-art methods in speech emotion recognition: CNN \cite{latif2019direct}, CNN-LSTM \cite{latif2019direct}, representation learning \cite{ghosh2016representation} and LSTM with Connectionist Temporal Modeling (LSTM-CTC) \cite{han2018towards}.

\subsubsection*{\bf Results} Table \ref{tab:all_baselines_speech} shows that our model performs better than the baselines and state-of-the-art methods on IEMOCAP. Our model's performance may seem only slightly better (1.3\%) compared to LSTM-CTC, but it requires 4 times more parameters than ours. LSTM-CTC uses 238-dimensional feature vectors where our feature dimension is only 35. Although PATCHY-Diff yields a competitive accuracy with a smaller model size on IEMOCAP, it trails L-GrIN by large margin on other databases. Note that PATCHY-SAN and PATCHY-Diff perform better than BLSTM and CNN-LSTM methods, indicating the effectiveness of graph-based methods in general. 
\subsection{Body emotion recognition}
\begin{figure}
    \centering
\begin{minipage}[t]{0.15\textwidth}
\centering
\includegraphics[width=1\linewidth, trim={0mm 5mm 0mm 2mm}, clip=true]{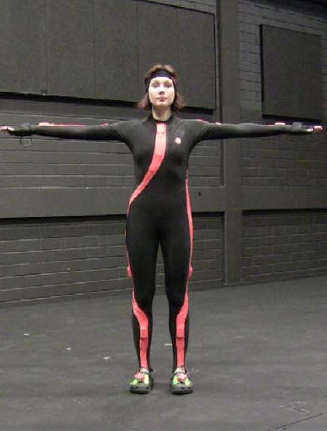}
\label{fig:IEMO_Adj}
\end{minipage}
\begin{minipage}[b]{0.15\textwidth}
\centering
\includegraphics[width=1\linewidth, trim={0mm 5mm 0mm 2mm}, clip=true]{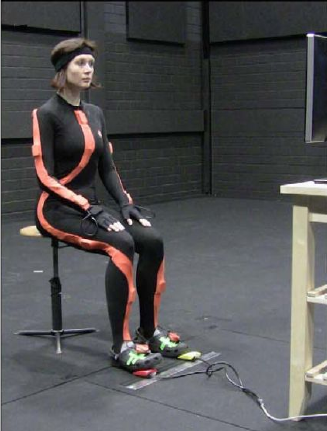}
\end{minipage}
\begin{minipage}[b]{0.15\textwidth}
\centering
\includegraphics[width=1\linewidth, trim={0mm 5mm 0mm 2mm}, clip=true]{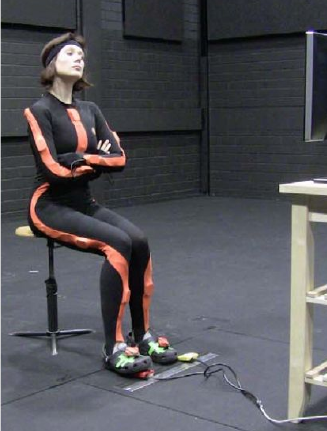}
\end{minipage}
\vspace{-3mm}
   \caption{Motion capture recording set-up for the MPI database showing an actor posing for (left to right) T pose (reference), neutral and pride pose.}
\label{fig:bodyDB}
\end{figure}
\begin{table}[t]
\caption{Body emotion recognition results on the MPI database.}
\label{tab:all_baselines_body}
\vspace{-3mm}
\begin{center}
\renewcommand*{\arraystretch}{1.5}
\begin{tabular}{p{3cm}| c | c}
\hline 
\bf \multirow{1}{*}{Model}  & \multicolumn{1}{c|}{\bf Accuracy (\%)} & \bf \multirow{1}{*}{Parameters}\\
 \hline \hline
$^*$BLSTM & 45.52            & $\sim 0.9$M \\
$^*$GCN    &   56.03     & $\sim 92$K \\
$^*$PATCHY-SAN \cite{niepert2016learning}  &  48.42 & $\sim 80$K\\ 
$^*$PATCHY-Diff \cite{ying2018hierarchical}  & 55.29  & $\sim 71$K \\
Trajectory learning \cite{crenn2017toward} & 50.00 & - \\
$^*${\bf L-GrIN} &  \bf{58.59} & $\sim 110$K \\
\hline
\multicolumn{3}{l}{$^*$ use same node features}
\end{tabular}
\end{center}
\end{table}
\subsubsection*{\bf Databases} 
We use the MPI emotional body expression database \cite{volkova2014mpi} for our experiments. This database contains 1447 body motion samples of actors narrating coherent stories labeled with 11 emotions: \emph{amusement, anger, disgust, fear, joy, neutral, pride, relief, sadness, shame,} and \emph{surprise}. During their performance, a mocap system (device model: Xsens MVN) recorded the human motion using miniature inertial sensors. The system recorded dynamic 3D postures from 22 joints with a sampling rate of $120$Hz. 
\subsubsection*{\bf Node features} For this database, we use the raw information provided by the mocap system. Each node contains the 3D positions and orientations (measure in terms of the Euler angles, pitch, yaw and roll) at a given time-step. These measurements come with the database. The feature consists of Euler angles from 22 joints and additional location information of the reference point. We use all the information (without any preprocessing) as node features, resulting into a vector of dimension $P = 72$.

\subsubsection*{\bf Implementation details} Each input sample produces a graph of $M = 120$ nodes, where each node corresponds to a temporal segment of 120$^{th}$ of a second. Cyclic padding is used as before. We perform a 5-fold cross-validation and report the average accuracy in Table \mbox{\ref{tab:all_baselines_body}}. All other network parameters remain the same as before.

\subsubsection*{\bf Baselines, state-of-the-art} 
Our model is compared with the baselines (BLSTM and GCN), the state-of-the-art graph-based architectures (PATCHY-SAN and PATCHY-Diff), and a recent work on this database, i.e., trajectory learning \cite{crenn2017toward}. The trajectory learning system \cite{crenn2017toward} models neural motion and analyzes the spectral difference between an expressive motion and a neutral motion in order to recognize the body expressions.

\subsubsection*{\bf Results} 
Table \ref{tab:all_baselines_body} shows that L-GrIN outperforms the baselines and state-of-the-art methods on the MPI body expression database. Graph-based methods continue to perform well, indicating the effectiveness of graph-based methods for such tasks. Fig.~\ref{fig:adj} shows the learned adjacency $\mathbf{A}$ for the MPI database. As before, the learned graph structure exhibit higher temporal dependencies among the neighboring nodes.
\subsection{Network analysis}
\begin{table}[t]
\caption{Comparison between learnable and fixed pooling strategies on the RML database. All experiments in this table use the same (binary) adjacency matrix for fair comparison.}
\vspace{-3mm}
 \label{tab:pool}
 \renewcommand*{\arraystretch}{1.3}
 \begin{center}
 \begin{tabular}{l c}
 \hline
 \bf Pooling  & \bf Accuracy $(\%)$ \\ \hline 
 Maxpool                                    & $89.76$      \\
 Meanpool                                   & $90.23$    \\
 $\mathrm{Sortpool}$ \cite{zhang2018end}     & $83.66$    \\
 Learnable pool      & $\mathbf{91.50}$    \\
\hline
 \end{tabular}
 \vspace{-4mm}
 \end{center}
 \end{table}
\begin{table}[t]
\caption{Comparison between learnable and manually constructed graph structures. For fair comparison, all experiments use maxpool to convert node embeddings to graph embeddings.}
\vspace{-4mm}
\label{tab:adj}
\renewcommand*{\arraystretch}{1.4}
 \begin{center}
 \begin{tabular}{l|ccc|ccc}
 \hline 
\multirow{2}{*}{}  & \multicolumn{3}{c|}{\bf Accuracy (\%)} & \multicolumn{3}{c}{\bf Params}\\
 & \scriptsize RML & \scriptsize IEMOCAP & \scriptsize MPI & \scriptsize RML & \scriptsize IEMOCAP & \scriptsize MPI \\
 \hline 
 Binary &  $89.5$ & $61.4$& $53.6$ & $113$K & $78$K &  $96$K \\
Weighted &  $62.4$ & $54.3$ & $49.0$  & $113$K & $78$K &  $96$K  \\
 Learnable & $\mathbf{91.5}$ & $\mathbf{65.5}$ & $\mathbf{58.9}$  & $120$K & $92$K & $110$K\\
\hline
 \end{tabular}
 \end{center}
 \end{table}
%
\subsubsection*{\bf Network size} Tables \ref{tab:all_baselines_faces}, \ref{tab:all_baselines_speech} and \ref{tab:all_baselines_body} list the number of learnable network parameters for the baselines, state-of-the-art graph-based architectures and the proposed L-GrIN. As mentioned earlier, a graph network largely reduces the number of learnable parameters as compared to the BLSTM or CNN architectures such as SENet (see Table \ref{tab:all_baselines_faces}) without compromising the recognition accuracy. Our model has more parameters than the baseline GCN due to the inception layers and other learnable parameters, but also improves the recognition accuracy significantly. PATCHY-SAN and PATCHY-Diff have smaller network size compared to L-GRIN, but both trail L-GrIN in terms of performance on all databases. In case of facial emotion recognition, we discount the model size of the landmark detector in the comparison as it is common to all except SENet. For speech and body emotion recognition, no additional network was required as we used hand-crafted features and raw data.

\vspace{2mm}
\subsubsection*{\bf Learnable vs.~fixed pooling} Recall that to obtain a graph-level embedding from node-level embeddings, L-GrIN learns a pooling function (see Fig.~\ref{fig:overview}). To show if learnable pooling indeed improves the recognition performance, we compare its performance with various fixed pooling strategies: max pooling, mean pooling and sort pooling ($\mathrm{sortpool}$) \cite{zhang2018end}. Table \ref{tab:pool} presents the comparisons on the RML database in terms of facial emotion recognition accuracy, which clearly shows the advantage of learnable pooling over fixed pooling strategies. Similar trend is observed for other databases.
\subsubsection*{\bf Learnable vs.\ manually constructed adjacency} An adjacency matrix represents the pairwise relationship between the graph nodes. When this information is not available naturally, a common practice is to manually construct an adjacency matrix. We argued earlier that this may result in sub-optimal graph structures which in turn affects the classification performance. We now compare the performance of leranable adjacency with two fixed adjacency matrices: \\(i) \textit{Binary adjacency:} a natural choice is a binary adjacency matrix as used for graph-based action recognition \cite{yan2018spatial}. This is defined as $(\mathbf{A}_b)_{ij} = 1$ if $\vert i -j \vert =1$ and $0$ otherwise, i.e., a node (frame) is connected only to its subsequent and preceding node in the temporal direction. \\(ii)~\textit{Weighted adjacency:} Another adjacency matrix is formed by using the squared $\ell_2$ distance between two node attributes as their edge weight. This is defined as $(\mathbf{A}_w)_{ij} = \|\mathbf{n}_i-\mathbf{n}_j\|_2^2$. 

Table \ref{tab:adj} compares the performance of the proposed learnable adjacency with the two fixed adjacency matrices described above on the RML, IEMOCAP and the MPI databases. We chose one database from every modality. For this set of experiments we used only maxpooling to obtain the graph-level embeddings for fair comparison. Clearly, the learnable adjacency matrix shows consistent improvement in accuracy across all databases for a relatively small increase in model complexity (only $6\%$ additional parameters). The results show that a learnable adjacency has better at generalizing across databases and modalities.
\begin{table}[tb]
\caption{Ablation study on the RML database. Each new component in L-GRIN contributes towards its performance.}
\label{tab:ablation}
\vspace{-1mm}
\renewcommand*{\arraystretch}{1.4}
\begin{tabular}{cccc c}
\hline
{\bf $\mathcal{G}^*$conv} & {\bf Inception} & \bf Learned $\mathbf{A}$  & \bf Learned $\mathbf{p}$ & {\bf Accuracy} (\%) \\ 
\hline \hline
- & - & -     &  - &    $76.57$       \\
\checkmark & - & -     &  - & $80.12$          \\
- & \checkmark & -     &  - & $87.58$          \\
- & - & \checkmark  &  - & $79.78$        \\
- & - & - & \checkmark  & $82.86$        \\
- & - & \checkmark  &    \checkmark & $84.21$        \\
\checkmark & \checkmark   &     -      &    -  & $90.65$        \\
\checkmark & \checkmark & \checkmark    &  - & $91.50$ \\
\checkmark & \checkmark & - & \checkmark     & $91.50$ \\
 \checkmark & \checkmark & \checkmark    & \checkmark &  $\mathbf{94.11}$\\
\hline
\end{tabular}
\end{table}
\begin{table}[tb]
\centering
\caption{Analyzing inception layer settings on the RML database.}
\vspace{-1mm}
\renewcommand*{\arraystretch}{1.5}
\begin{tabular}{c|c}
\hline 
\multicolumn{2}{c}{ \it Effect of filter size ($\eta$)}\\
\hline
\bf Size of the two filters & \bf Accuracy (\%) \\ \hline 
 $(16, 32)$     & $90.82$      \\
$(32, 64)$    & $92.47$      \\ 
$\mathbf{(64, 128)}$    & $\mathbf{94.11}$    \\
$(128, 256)$    & $93.13$     \\ \hline
\multicolumn{2}{c}{\it Effect of number of inception layers }\\ \hline
\bf Number of layers & \bf Accuracy (\%)\\
\hline
 $1$ & $91.77$ \\
 $\mathbf{2}$ & $\mathbf{94.11}$ \\
 $ 3 $  &  $90.78$ \\
 \hline
\end{tabular}
\label{tab:ablation_filter}
\end{table}
%
%
\begin{figure}[tb]
 \centering
 \begin{minipage}[t]{0.7\linewidth}
    \centering
  \includegraphics[width=1\linewidth, trim={7mm 0mm 5mm 5mm}, clip= true]{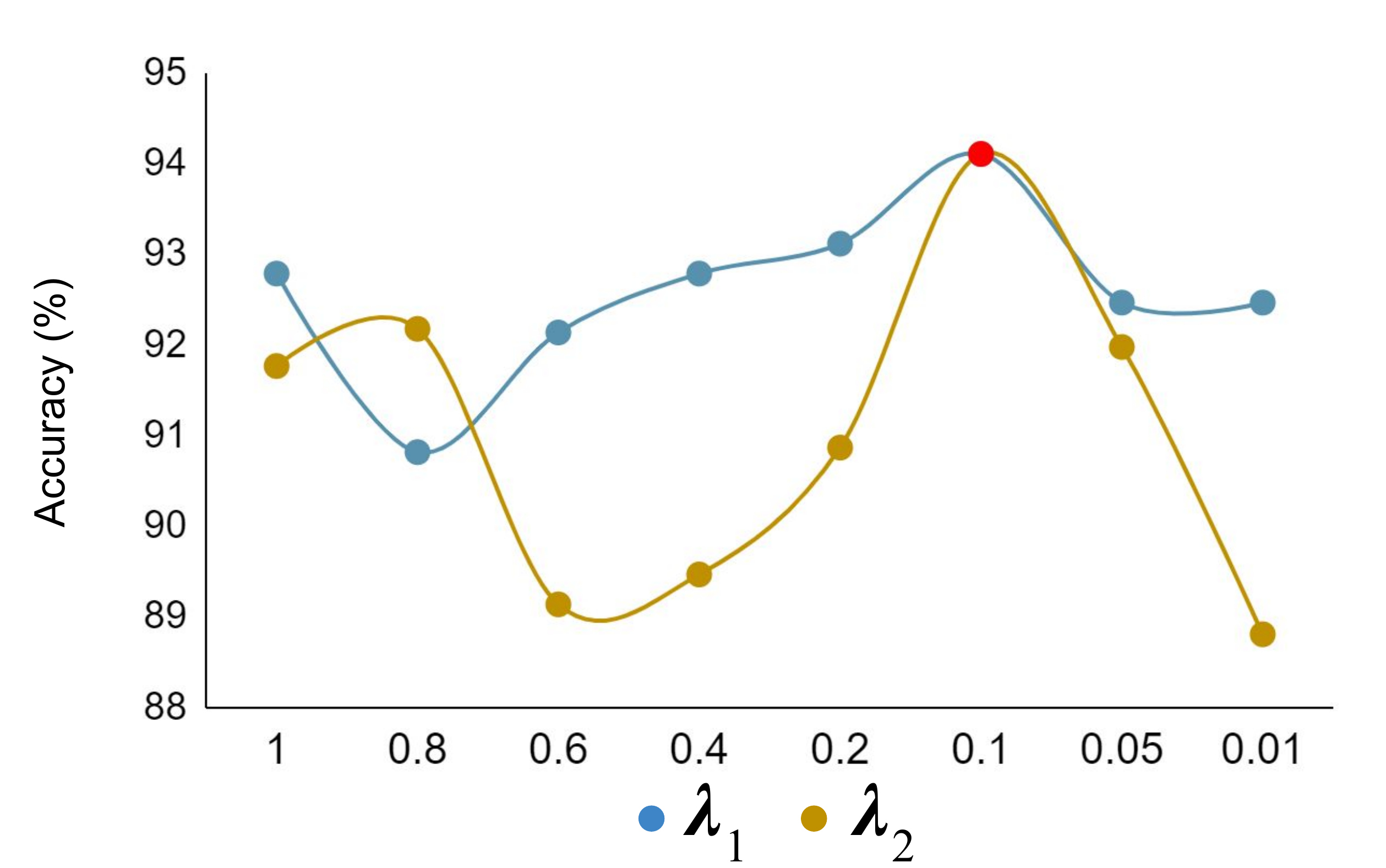}
 \end{minipage}\hspace{1mm}
 \begin{minipage}[t]{0.7\linewidth}
    \centering
    \includegraphics[width=1\linewidth, trim={7mm 2mm 5mm 7mm}, clip= true]{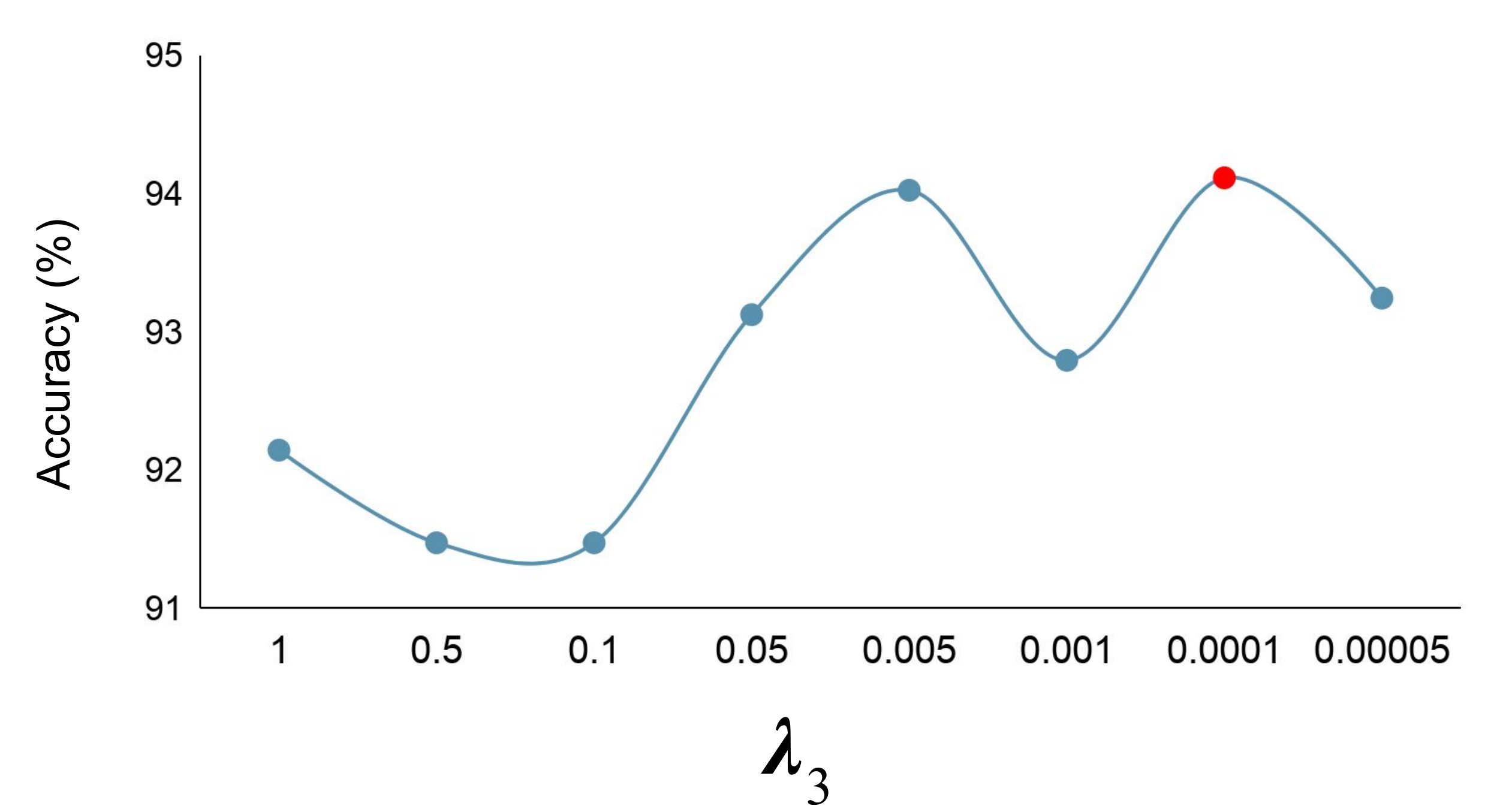}
 \end{minipage}
 \vspace{-0.1cm}
 \caption{Effect of the weight parameters in the loss function; experiments on the RML database.}
 \label{fig:lambda}
\end{figure}
\subsubsection*{\bf Ablation study} \label{subsubsec:ablation}
We performed exhaustive ablation experiments to investigate the contribution of each component we proposed to build L-GrIN. Table~\ref{tab:ablation} presents the ablation results on the RML database. We observe that each new component brings significant improvement (row 2 to row 5) over the performance of standard GCN \cite{kipf2017semi} which has $76.57\%$ recognition accuracy (the top row in Table~\ref{tab:ablation}). The introduction of the graph inception layer increases the recognition rate by $11\%$; when combined with our new graph convolution layer $\mathcal{G}^*$conv (Eq.~\eqref{eq:gcn_MLP}), the accuracy increases to $90.65\%$. Adding the learnable graph structure (learned $\mathbf{A}$) and learnable pooling bring the accuracy up to $94.11\%$ both contributing to the accuracy. Removing either of the leanrable components reduces the accuracy by $2.61\%$. 
The ablation results show that each of the proposed components in our architecture is important, and contributes positively towards its superior performance. Similar ablation trend was observed for other databases.

\subsubsection*{\bf Inception layer settings}
We also investigate the effects of the graph inception layer hyperparameters: (i) the parameter $\eta$ corresponding to the size of the graph convolution filters $\mathcal{G}_1^*$ and $\mathcal{G}_2^*$ in Eq.~\mbox{\eqref{eq:incept}}, and (ii) the number of graph inception layers in L-GrIN. First, we vary the filter dimensions (can be interpreted as scales) in the two inception layers and note how this correspond to the model's performance. Results for the RML database is presented in Table \ref{tab:ablation_filter}; similar trends have been observed for other databases. Results in Table \ref{tab:ablation_filter} show that we achieve the best performance for the combination of (64, 128), which is used in our model. Next, we vary the number of inception layers in the model, each with (64, 128) filter combination (see Table \ref{tab:ablation_filter}. We observe that reducing or increasing the number of inception layers from 2 results in a drop in performance. We chose to use two inception layers in the proposed model. It is obvious that the model size increases significantly as we add more inception layers or increase filter sizes within the layers. We notice a small drop in performance with larger filter sizes and with higher number of inception layers. This could be possibly due to over-smoothing and over-mixing of the node features. However, the over-smoothing effect is not as prominent as in many node classification tasks.
\subsubsection*{\bf Analysis of the control weights} We also examine the impact of the weights controlling the various components of the loss function in Eq.~\eqref{eq:GL-loss}, i.e., $\lambda_1$, $\lambda_2$ and $\lambda_3$. Fig.~\mbox{\ref{fig:lambda}} shows that highest performance is achieved for $\lambda_1= \lambda_2 = 0.1$ and $\lambda_2=0.0001$ (marked red in the plots) on the RML database. We use these $\lambda$ values in our experiments.
\subsubsection*{\bf Cross-corpus performance} Methods exhibiting superior performance on one corpus, often fall short when tested on another corpus having different statistical distributions. We investigated the ability of our model to generalize across databases by evaluating its cross-corpus performance. To this end, we trained L-GrIN on one database, followed by fine-tuning a fully-connected layer on the target database, without changing the graph structure (or other parameters) learned from the training database.

Results in Table~\ref{tab:cross} shows that our model can generalize well producing consistent results under cross-corpus evaluation. Our cross-corpus results higher accuracy compared to the same-corpus GCN accuracy. Cross-corpus results are comparable with the same-corpus performance of PATCHY-SAN. This shows the strength of the proposed architecture. It is worth noticing that the RML database (when used for training) does not have \textit{neutral} and \textit{calmness} emotion classes, but our model still recognizes those emotions on RAVDESS with $67.2\%$ and $73.4\%$ accuracy.
\begin{table}[tb]
\centering
\caption{Cross-corpus performance of our model (L-GrIN) for facial emotion recognition.}
\vspace{-1mm}
\renewcommand*{\arraystretch}{1.5}
\begin{tabular}{c|c|c}
\hline 
\bf Trained on  & \bf Evaluated on & \bf Accuracy (\%) \\ \hline \hline
\multirow{2}{*}{RAVDESS}  & RML & $81.94$ \\ 
        & eNTERFACE          &    $75.80$       \\ 
  \hline
\multirow{2}{*}{RML}   & RAVDESS & $75.42$\\
   & eNTERFACE  &  $61.71$       \\ \hline
\multirow{2}{*}{eNTERFACE}      & RML & $79.86$ \\
         & RAVDESS & $77.51$\\ \hline
\end{tabular}
\label{tab:cross}
\end{table}
\begin{figure*}[t]
\begin{center}
\includegraphics[width=0.9\linewidth,trim={0cm 0cm 0cm 0cm, clip}, clip=true]{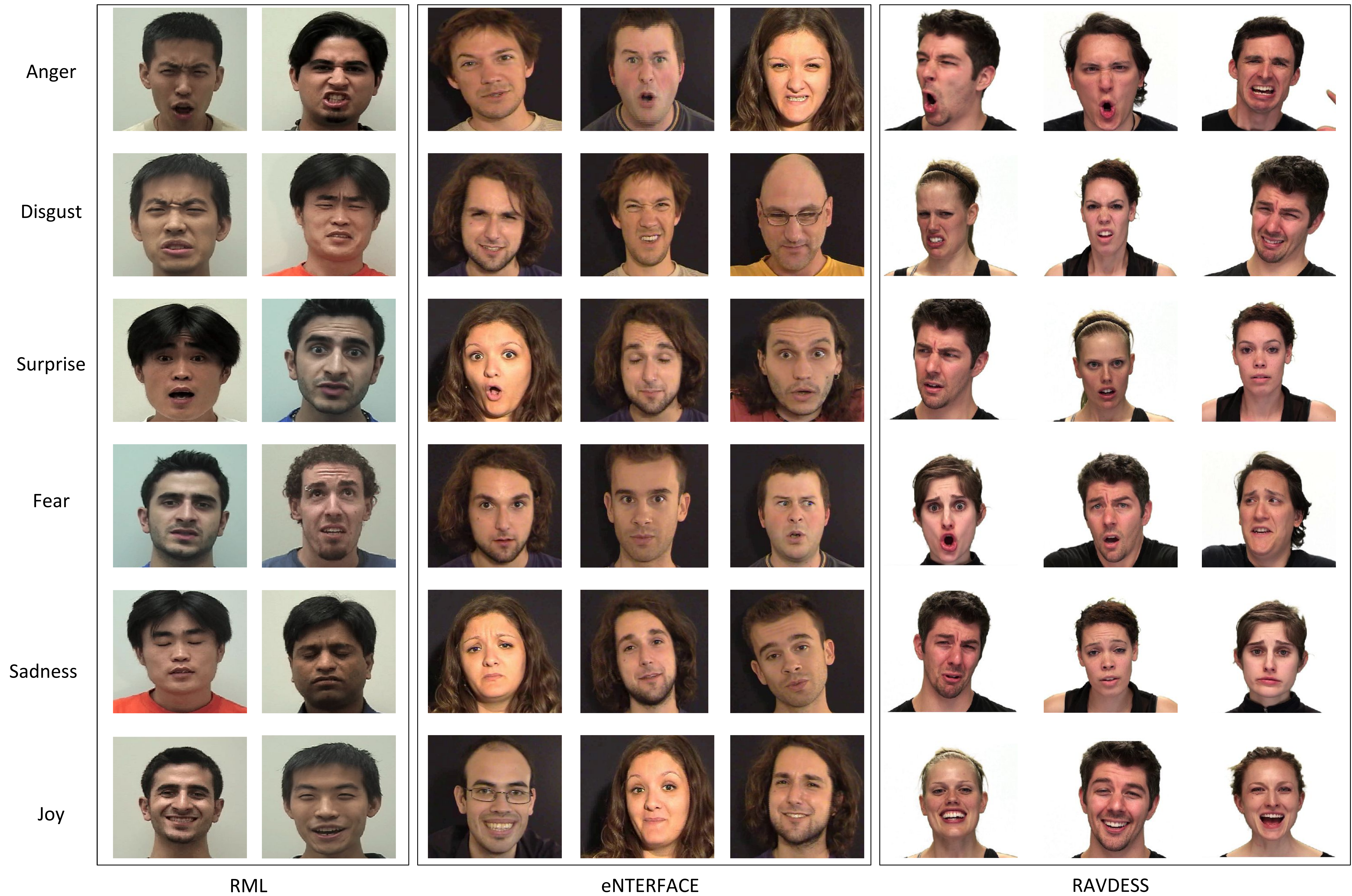}
\end{center}
\vspace{-0.2cm}
   \caption{Qualitative results showing the node (frame) for a graph input that generated the strongest response in our network. One result is displayed per class for the three databases. This shows that L-GRIN is able to learn the salient information for each emotion.}
\label{fig:qual}
\end{figure*}
\subsubsection*{\bf Network visualization} 
To get an insight into the learning process of our model, we visualized how it attends to different nodes. The video data are the most suitable for the visualization. We use our trained model, and then feed-forward each test video sample through the network, and identify the node (each node corresponds to a video frame) that responded most strongly towards the maxpooling layer. This yields a \emph{salient} node corresponding to each input. We present the corresponding video frames - one example per emotion class for RML, eNTERFACE and RAVDESS databases in Fig.~\ref{fig:qual}. The results show that the proposed model is able to learn the salient information from the input graphs such that it is representative of each emotion.

%
\section{Conclusion}
\label{sec:conclusions}
We proposed a novel, generalized graph architecture that can recognize emotion in a variety of dynamic input sequence. Our proposed architecture, {L-GrIN}, learns to detect emotion while jointly learning the underlying graph structure (adjacency matrix) and a pooling function to yield graph-level representation from node-level embeddings. We proposed a new spectral graph convolution operation and introduced the idea of inception in the graph domain. 
The advantage of our model lies in its state-of-the-art performance spanning three different modalities (video, audio and motion capture), with significantly fewer parameters compared to the CNNs and RNNs. This indicates that our model is suitable for applications in resource-constrained devices, such as smartphones.

We used both modality-specific features and even raw data as node features in this work. Our approach is not tied to any particular feature. In fact, our model can be trained end-to-end by combining it with modality-specific networks (e.g., a CNN) for feature extraction. The architecture we developed, although focuses on emotion recognition, is fairly generic. It will be applicable to a variety of classification tasks involving dynamic data, such as pose estimation, action recognition and visual speech recognition. Since our model makes no assumption about the graph structure, this is also applicable to common unstructured graphs. Future work will be directed towards building multimodal graph architectures taking advantage of the modality-agnostic architecture.
\balance
\bibliographystyle{IEEEtran}
\bibliography{egbib}
\end{document}